\pdfoutput=1

\documentclass[11pt]{article}

\usepackage[]{acl}

\usepackage{times}
\usepackage{latexsym}
\usepackage{graphicx}
\usepackage{algorithm}
\usepackage{algorithmic}
\usepackage{amsmath,amsthm,mathtools}
\usepackage{booktabs}
\usepackage{multirow}
\usepackage{xcolor}
\usepackage{subfig}
\usepackage{MnSymbol}

\usepackage[T1]{fontenc}

\usepackage[utf8]{inputenc}

\usepackage{microtype}

\title{Schema-Guided User Satisfaction Modeling for Task-Oriented Dialogues}

\author{Yue Feng $^\dagger$\thanks{\, Work done while Yue Feng was an intern at Amazon, Alexa Shopping.} \quad Yunlong Jiao $^\ddagger$ \quad Animesh Prasad $^\ddagger$\\ \textbf{Nikolaos Aletras} $^\diamond$$^\ddagger$ \quad \textbf{Emine Yilmaz} $^\dagger$$^\ddagger$ \quad \textbf{Gabriella Kazai} $^\ddagger$ \\ 
  $^\dagger$University College London, London, UK\\
  $^\ddagger$Amazon, London, United Kingdom\\
  $^\diamond$University of Sheffield, Sheffield, UK\\ 
  {$^\dagger$\texttt{\{yue.feng.20,emine.yilmaz\}@ucl.ac.uk}} \\ {$^\ddagger$\texttt{\{jyunlong,gkazai\}@amazon.co.uk}}\\ {$^\diamond$\texttt{n.aletras@sheffield.ac.uk}}}

\begin{document}
\maketitle

\begin{abstract}
User Satisfaction Modeling (USM) is one of the popular choices for task-oriented dialogue systems evaluation, where
user satisfaction typically depends on whether the user's task goals were fulfilled by the system. Task-oriented dialogue systems use task schema, which is a set of task attributes, to encode the user's task goals. 
Existing studies on USM neglect explicitly modeling the user’s task goals fulfillment using the task schema. 
In this paper, we propose SG-USM, a novel schema-guided user satisfaction modeling framework. It explicitly models the degree to which the user's preferences regarding the task attributes are fulfilled by the system for predicting the user's satisfaction level. 
SG-USM employs a pre-trained language model for encoding dialogue context and task attributes. Further, it employs a fulfillment representation layer for learning how many task attributes have been fulfilled in the dialogue, an importance predictor component for calculating the importance of task attributes. Finally, it predicts the user satisfaction based on task attribute fulfillment and task attribute importance. Experimental results on benchmark datasets (i.e. MWOZ, SGD, ReDial, and JDDC) show that SG-USM consistently outperforms competitive existing methods. Our extensive analysis demonstrates that SG-USM can improve the interpretability of user satisfaction modeling, has good scalability as it can effectively deal with unseen tasks and can also effectively work in low-resource settings by leveraging unlabeled data.\footnote{Code is available at \url{https://github.com/amzn/user-satisfaction-modeling}.}

\end{abstract}

\begin{figure}[!t]
\centering
\includegraphics[scale=0.55]{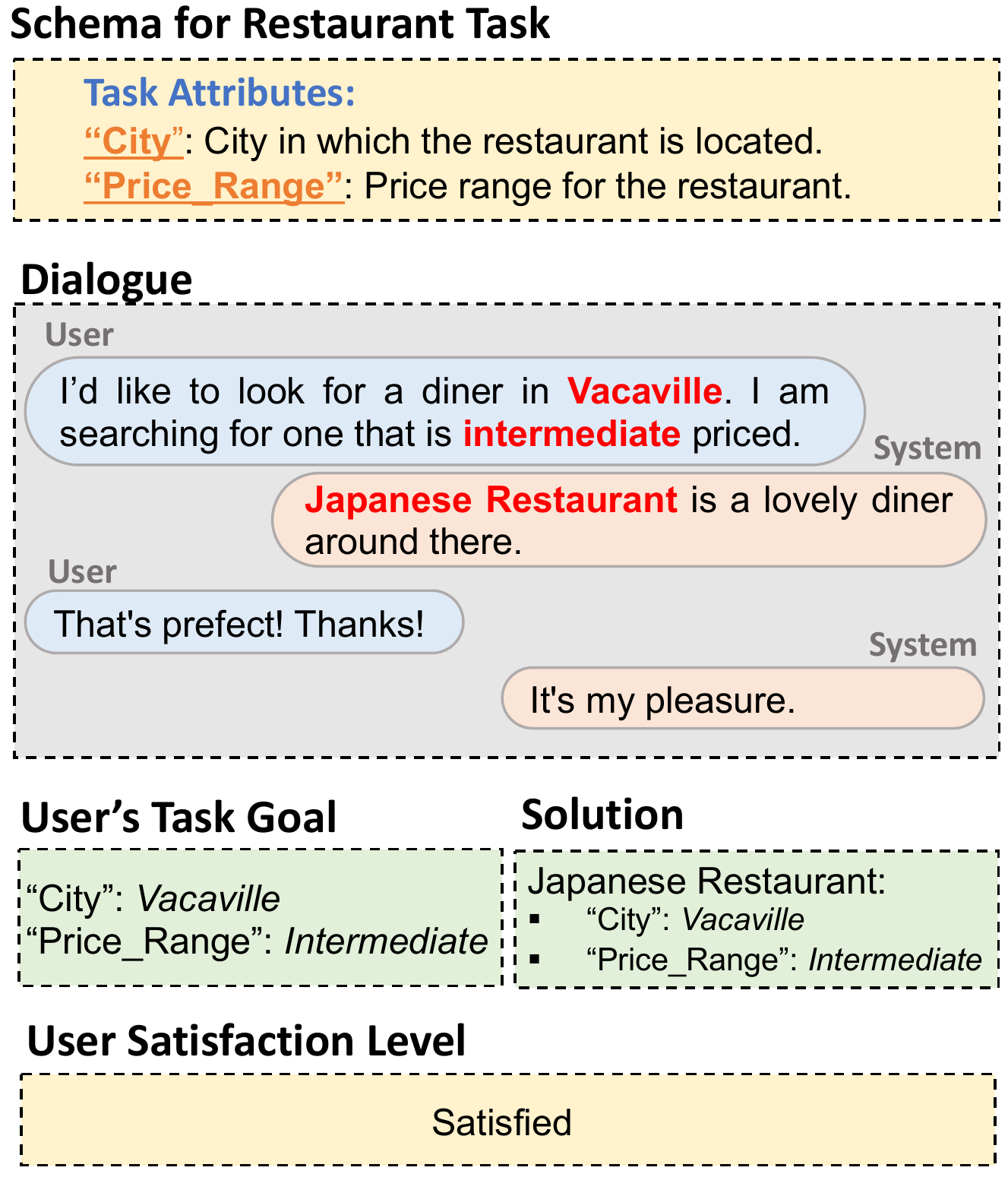}
\caption{Task-oriented dialogue system has a predefined schema for each task, which is composed of a set of task attributes. In a dialogue, the user's task goal is encoded by the task attribute and value pairs. The user is satisfied with the service when the provided solution fulfills the user's preferences for the task attributes.}
\label{fig:example}
\end{figure}

\section{Introduction}
Task-oriented dialogue systems have emerged for helping users to solve specific tasks efficiently~\citep{hosseini2020simple}.
Evaluation is a crucial part of the development process of such systems. Many of the standard automatic evaluation metrics, e.g. BLEU~\citep{papineni2002bleu}, ROUGE~\citep{lin2004rouge}, have been shown to be ineffective in task-oriented dialogue evaluation~\citep{deriu2021survey,liu2016not}. As a consequence, User Satisfaction Modeling (USM)~\citep{sun2021simulating,kachuee2021self,bodigutla2020joint,song2019using,rebensburgautomatic} has gained momentum as the core evaluation metric for task-oriented dialogue systems. 
USM estimates the overall satisfaction of a user interaction with the system. In task-oriented dialogue systems, whether a user is satisfied largely depends on how well the user’s task goals were fulfilled. Each task would typically have an associated task schema, which is a set of task attributes (e.g. location, date for check-in and check-out, etc. for a hotel booking task), and for the user to be satisfied, the system is expected to fulfill the user's preferences about these task attributes.  Figure~\ref{fig:example} shows an example of USM for task-oriented dialogues.

Effective USM models should have the following abilities: 
(1) Interpretability by giving insights on what aspect of the task the system performs well. For instance, this can help the system to recover from an error and optimize it toward an individual aspect to avoid dissatisfaction. 
(2) Scalability in dealing with unseen tasks, e.g. the model does not need to retrain when integrating new tasks.
(3) Cost-efficiency for performing well in low-resource settings where it is often hard to collect and expensive to annotate task-specific data. 

Previous work in USM follows two main lines of research. First, several methods use user behavior or system actions to model user satisfaction. In this setting, it is assumed that user satisfaction can be reflected by user behaviors or system actions in task-oriented dialogue systems, such as click, pause, request, inform~\citep{deng2022user,guo2020deep}. A second approach is to analyze semantic information in user natural language feedback to estimate user satisfaction, such as sentiment analysis~\citep{sun2021simulating,song2019using} or response quality assessment~\citep{bodigutla2020joint,zeng2020overview}.
However, both of these two lines of work do not take into account the abilities of interpretability, scalability, and cost-efficiency.

In this paper, we propose a novel approach to USM, referred to as Schema-Guided User Satisfaction Modeling (SG-USM). We hypothesize that user satisfaction should be predicted by the fulfillment degree of the user’s task goals that are typically represented by a set of task attribute and value pairs. Therefore, we explicitly formalize this by predicting how many task attributes fulfill the user's preferences and how important these attributes are. When more important attributes are fulfilled, task-oriented dialogue systems should achieve better user satisfaction. 

Specifically, SG-USM comprises a pre-trained text encoder to represent dialogue context and task attributes, a task attribute fulfillment representation layer to represent the fulfillment based on the relation between the dialogue context and task attributions, a task attribute importance predictor to calculate the importance based on the task attributes popularity in labeled and unlabeled dialogue corpus, and a user satisfaction predictor which uses task attributes fulfillment and task attributes importance to predict user satisfaction. 
SG-USM  uses task attributes fulfillment and task attributes importance to explicitly model the fulfillment degree of the user's task goals (interpetability). It uses an task-agnostic text encoder to create representations of task attributes by description, no matter whether the task are seen or not (scalability). Finally, it uses unlabeled dialogues in low-resource settings (cost-efficiency). 

Experimental results on popular task-oriented benchmark datasets show that SG-SUM substantially and consistently outperforms existing methods on user satisfaction modeling. Extensive analysis also reveals the significance of explicitly modeling the fulfillment degree of the user's task goals, the ability to deal with unseen tasks, and the effectiveness of utilizing unlabeled dialogues.


\section{Related Work}

\paragraph{Task-oriented Dialogue Systems.}
Unlike chitchat dialogue systems that aim at conversing with users without specific goals, task-oriented dialogue systems assist users to accomplish certain tasks~\citep{feng2021sequence,eric2020multiwoz}. Task-oriented dialogue systems can be divided into module-based methods~\citep{feng2022dynamic,ye2022assist,su2022multi,heck2020trippy,chen2020schema,wu2019transferable,lei2018sequicity,liu2016attention} and end-to-end methods~\citep{feng2022topic,qin2020dynamic,yang2020graphdialog,madotto2018mem2seq,yao2014recurrent}. 
To measure the effectiveness of task-oriented dialogue systems, evaluation is a crucial part of the development process. Several approaches have been proposed including automatic evaluation metrics~\citep{rastogi2020towards,mrkvsic2017neural}, human evaluation~\citep{feng2022topic,goo2018slot}, and user satisfaction modeling~\citep{sun2021simulating,mehrotra2019jointly}. Automatic evaluation metrics, such as BLEU~\citep{papineni2002bleu}, make a strong assumption for dialogue systems, which is that valid responses have significant word overlap with the ground truth responses. However, there is significant diversity in the space of valid responses to a given context~\citep{liu2016not}. Human evaluation is considered to reflect the overall performance of the system in a real-world scenario, but it is intrusive, time-intensive, and does not scale~\citep{deriu2021survey}. Recently, user satisfaction modeling has been proposed as the main evaluation metric for task-oriented dialogue systems, which can address the issues listed above. 

\begin{figure*}[!t]
\centering
\includegraphics[scale=0.5]{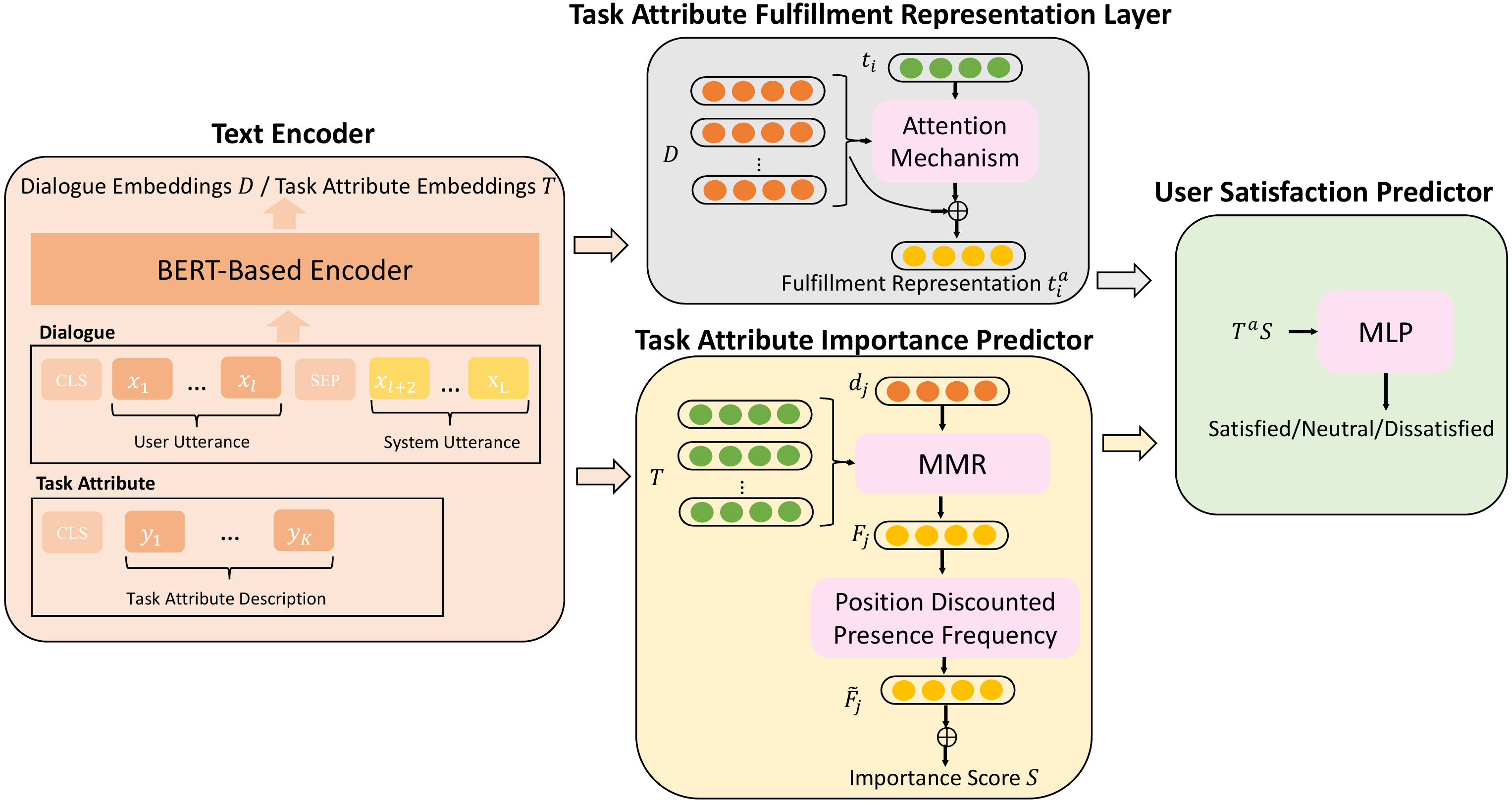}
\caption{The architecture of SG-USM for user satisfaction modeling on task-oriented dialogues.}
\label{fig:framework}
\end{figure*}

\paragraph{User Satisfaction Modeling.}
User satisfaction in task-oriented dialogue systems is related to whether or not, or to what degree, the user's task goals are fulfilled by the system. Some researchers study user satisfaction from temporal user behaviors, such as click, pause, etc.~\citep{deng2022user, guo2020deep, mehrotra2019jointly, wu2019influence, su2018user,mehrotra2017user}. Other related studies view dialogue action recognition as an important preceding step to USM, such as request, inform, etc.~\citep{deng2022user,kim2022multi}. However, sometimes the user behavior or system actions are hidden in the user’s natural language feedback and the system's natural language response~\citep{hashemi2018measuring}. To cope with this problem, a number of methods are developed from the perspective of sentiment analysis~\citep{sun2021simulating,song2019using, engelbrecht2009modeling} and response quality assessment~\citep{bodigutla2020joint,zeng2020overview}. However, all existing methods cannot explicitly predict user satisfaction with fine-grained explanations, deal with unseen tasks, and alleviate low-resource learning problem. Our work is proposed to solve these issues.

\section{Schema-guided User Satisfaction Modeling}
Our SG-USM approach formalizes user satisfaction modeling by representing the user's task goals as a set of task attributes, as shown in Figure~\ref{fig:example}. The goal is to explicitly model the degree to which task attributes are fulfilled, taking into account the importance of the attributes. As shown in Figure~\ref{fig:framework}, SG-USM consists of a text encoder, a task attribute fulfillment representation layer, a task attribute importance predictor, and a user satisfaction predictor. Specifically, the text encoder transforms dialogue context and task attributes into dialogue embeddings and task attribute embeddings using BERT~\citep{kenton2019bert}. The task attribute fulfillment representation layer models relations between the dialogue embeddings and the task attribute embeddings by attention mechanism to create task attribute fulfillment representations. Further, the task attribute importance predictor models the task attribute popularity in labeled and unlabeled dialogues by the ranking model to obtain task attribute importance weights. Finally, the user satisfaction predictor predicts user satisfaction score on the basis of the task attribute fulfillment representations and task attribute importance weights using a multilayer perceptron.


\subsection{Text Encoder}
The text encoder takes the dialogue context (user and system utterances) and the descriptions of task attributes as input and uses BERT to obtain dialogue and task attribute embeddings, respectively. 

Considering the limitation of the maximum input sequence length of BERT, we encode dialogue context by each dialogue turn. Specifically, the BERT encoder takes as input a sequence of tokens with length $L$, denoted as $X = (x_1, ..., x_L )$. The first token $x_1$ is [CLS], followed by the tokens of the user utterance and the tokens of the system utterance in one dialogue turn, separated by [SEP]. The representation of [CLS] is used as the embedding of the dialogue turn. Given a dialogue with $N$ dialogue turns, the output dialogue embeddings is the concatenation of all dialogue turn embeddings $D = [d_1;d_2; ...; d_N]$.

To obtain task attribute embeddings, the input is a sequence of tokens with length $K$, denoted as $Y = \{y_1, ..., y_K\}$. The sequence starts with [CLS], followed by the tokens of the task attribute description. The representation of [CLS] is used as the embedding of the task attribute. The set of task attribute embeddings are denoted as $T = \{t_1, t_2,...,t_M\}$, where $M$ is the number of task attributes.

\subsection{Task Attribute Fulfillment Representation Layer}
The task attribute fulfillment representation layer takes the dialogue and task attribute embeddings as input and calculates dialogue-attended task attribute fulfillment representations. This way, whether each task attribute can be fulfilled in the dialogue context is represented. 

Specifically, the task attribute fulfillment representation layer constructs an attention vector by a bilinear interaction, indicating the relevance between dialogue and task attribute embeddings. Given the dialogue embeddings $D$ and $i$-th task attribute embedding $t_i$ , it calculates the relevance as follows,
\begin{gather}
A_i = \text{softmax}(\text{exp}(D^\mathsf{T} W_a t_i)),
\end{gather}
where $W_a$ is the bilinear interaction matrix to be learned. $A_i$ represents the attention weights of
dialogue turns with respect to the $i$-th task attribute. Then the dialogue-attended $i$-th task attribute fulfillment representations are calculated as follows,
\begin{gather}
t_i^a = DA_i.
\end{gather}
The dialogue-attended task attribute fulfillment representations for all task attributes are denoted as:
\begin{gather}
T^a = [t_1^a, t_2^a,...,t_M^a].
\end{gather}
where $M$ is the number of the task attributes.

\subsection{Task Attribute Importance Predictor}
The task attribute importance predictor also takes the dialogue and task attribute embeddings as input and calculates attribute importance scores. The importance scores  are obtained by considering both the task attribute presence frequency and task attribute presence position in the dialogue. 

First, we use the Maximal Marginal Relevance (MMR)~\citep{carbonell1998use} to select the top relevant task attributes for the dialogue context. The selected task attributes are then used to calculate the task attribute presence frequency in the dialogue. The MMR takes the $j$-th dialogue turn embeddings $d_j$ and task attribute embeddings $T$ as input, and picks the top $K$ relevant task attributes for the $j$-th dialogue turn:
\begin{equation}
\small
R_j = \operatorname*{argmax}_{t_i \in T \setminus U} [\lambda \text{cos}(t_i, d_j) - (1-\lambda)\operatorname*{max}_{t_k \in U} \text{cos}(t_i, t_k)]
\end{equation}
where $U$ is the subset of attributes already selected as top relevant task attributes, $\text{cos}()$ is the cosine similarity between the embeddings. $\lambda$ trades off between the similarity of the selected task attributes to the dialogue turn and also controls the diversity among the selected task attributes. The task attribute presence frequency vector for the $j$-th dialogue turn is computed as follows,
\begin{gather}
F_j = [f_j^1, f_j^2, f_j^3, ..., f_j^M] \\
f_j^i = 
\begin{cases}
1 &  i \in R_j \\
0 &  i \notin R_j
\end{cases}
\end{gather}
where $M$ is the number of the task attributes.

However, the task attribute presence frequency vector does not reward task attributes that appear in the beginning of the dialogue. The premise of task attribute importance score is that task attributes appearing near the end of the dialogue should be penalized as the graded importance value is reduced logarithmically proportional to the position of the dialogue turn. A common effective discounting method is to divide by the natural log of the position:
\begin{gather}
\widetilde{F_j} = \frac{F_j}{log(j+1)}
\end{gather}
The task attribute importance predictor then computes the importance score on the basis of the sum of the discounted task attribute presence frequency of all dialogues. Given the dialogue corpus (including both labeled and unlabeled dialogues) with Z dialogues $C=\{D_1, D_2, ..., D_Z\}$, the task attribute importance scores are calculated as follow:
\begin{gather}
S = \text{softmax}(\sum_{l= 1}^Z\sum_{j=1}^{\text{Num}(D_l)}\widetilde{F_j^l})
\end{gather}
where $\text{Num}()$ is the number of the dialogue turn in dialogue $D_l$, and $\widetilde{F_j^l}$ is the discounted task attribute presence frequency of $j$-th dialogue turn in dialogue $D_l$.

\subsection{User Satisfaction Predictor}
Given the dialogue-attended task attribute fulfillment representations $T^a$ and the task attribute importance scores $S$, the user satisfaction labels are obtained by aggregating task attribute fulfillment representations based on task attribute importance scores. This way, the user satisfaction is explicitly modeled by the fulfillment of the task attributes and their individual importance.

Specifically, an aggregation layer integrates the dialogue-attended task attribute fulfillment representations by the task attribute importance scores as follows:
\begin{gather}
h = T^a S
\end{gather}
Then the Multilayer Perceptron (MLP)~\citep{hastie2009elements} with softmax normalization is employed to calculate the probability distribution of user satisfaction classes:
\begin{gather}
p = \text{softmax}(\text{MLP}(h))
\end{gather}

\subsection{Training}
We train SG-USM in an end-to-end fashion by minimizing the cross-entropy loss between the predicted user satisfaction probabilities and the ground-truth satisfaction:
\begin{gather}
\mathcal{L} = -y\text{log}(p)
\end{gather}
where $y$ is the ground-truth user satisfaction. Pre-trained BERT encoders are used for encoding representations of utterances and schema descriptions respectively. The encoders are fine-tuned during the training process.

\section{Experimental Setup}
\subsection{Datasets}
We conduct experiments using four benchmark datasets containing task-oriented dialogue on different domains and languages (English and Chinese), including MultiWOZ2.1 (MWOZ)~\citep{eric2020multiwoz}, Schema Guided Dialogue (SGD)~\citep{rastogi2020towards}, ReDial~\citep{li2018towards}, and JDDC~\citep{chen2020jddc}. 

\noindent\textbf{MWOZ and SGD} are English multi-domain task-oriented dialogue datasets, which include hotel, restaurant, flight, etc. These datasets contain domain-slot pairs, where the slot information could correspond to the task attributes.

\noindent\textbf{ReDial} is an English conversational recommendation dataset for movie recommendation. The task attributes are obtained from the Movie\footnote{\href{https://schema.org/Movie}{https://schema.org/Movie}} type on Schema.org.

\noindent\textbf{JDDC} is a Chinese customer service dialogue dataset in E-Commerce. The task attributes are obtained from the Product\footnote{\href{https://schema.org.cn/Product}{https://schema.org.cn/Product}} type on Schema.org.cn, which provides schemas in Chinese.

Specifically, we use the subsets of these datasets with the user satisfaction annotation for evaluation, which is provided by Sun et al~\citep{sun2021simulating}. We also use the subsets of these datasets without the user satisfaction annotation to investigate the semi-supervised learning abilities of SG-USM. Table~\ref{tab:datasets} displays the statistics of the datasets in the experiments.

\begin{table}[h]
\centering

\resizebox{0.49\textwidth}{!}{%
\begin{tabular}{l|cccc}
        \toprule
        \bf{Characteristics}&\bf{MWOZ}&\bf{SGD} &\bf{ReDial} &\bf{JDDC} \\
 		\hline
        \hline
        \text{Language} & English & English & English & Chinese \\
        \text{\#Dialogues} & 1,000 & 1,000 & 1,000 & 3,300\\
        \text{\#Utterances} & 12,553 & 13,833& 11,806& 54,517\\
        \text{\#Avg Turn} &  23.1 &  26.7&  22.5&32.3\\
        \text{\#Attributes} & 37 & 215&  128&13\\
        \text{\%Sat. Class} & 27:39:34 & 22:30:48&  23:26:51&23:53:24\\
        \hline
		\text{\#TrainSplit} & 7,648 &8,674 & 7,372 & 38,146\\
		\text{\#ValidSplit} & 952 &1,074 & 700&5,006\\
		\text{\#TestSplit} & 953 &1,085 & 547&4,765\\
		\text{\#Unlabeled Dialogues} & 4,000 &4,000 & 4,000&4,000\\
		\bottomrule
	\end{tabular}
	}
\caption{Statistics of the task-oriented dialogue datasets.} 
\label{tab:datasets}
\end{table}

\begin{table*}[!h]
\centering
\resizebox{1.0\textwidth}{!}{
\begin{tabular}{l|cccc|cccc|cccc|cccc}
        \toprule
        \multirow{2}{*}{{ \bf{Model}}}&\multicolumn{4}{c|}{{ \bf{MWOZ}}}&\multicolumn{4}{c|}{{ \bf{SGD}}}&\multicolumn{4}{c|}{{ \bf{ReDial}}}&\multicolumn{4}{c}{{ \bf{JDDC}}}\\
        \cline{2-17}
        &{Acc} &{P} &{R} &{F1} &{Acc} &{P} &{R} &{F1} &{Acc} &{P} &{R} &{F1} &{Acc} &{P} &{R} &{F1}\\
 		\hline
        \hline
        \text{HiGRU} & 44.6 &43.7 &44.3 &43.7 &50.0 &47.3 &48.4 &47.5 &46.1 &44.4 &44.0 &43.5 & 59.7 &57.3 &50.4 &52.0\\
        \text{HAN} & 39.0 &37.1 &37.1 &36.8 & 47.7 &47.1 &44.8 &44.9 & 46.3 &40.0 &40.3 &40.0 & 58.4 &54.2 &50.1 &51.2\\
        \text{Transformer} &42.8&41.5&41.9& 41.7 &53.1&48.3&49.9& 49.1 &47.5&44.9&44.7& 44.8 &60.9&59.2&53.4& 56.2\\
        \text{BERT} & 46.1 &45.5 &47.4 &45.9 & 56.2 &55.0 &53.7 &53.7 & 53.6 &50.5 &51.3 &50.0 & 60.4 &59.8 &58.8 &59.5\\
        \text{USDA} & \underline{49.9} & \underline{49.2} & \underline{49.0} & \underline{48.9} & \underline{61.4} & \underline{60.1} & \underline{55.7} & \underline{57.0} & \underline{57.3} & \underline{54.3} & \underline{52.9} & \underline{53.4} & \underline{61.8} & \underline{62.8} & \underline{63.7} & \underline{61.7}\\
        \hline
        \text{SG-USM-L} &50.8$^*$& 49.3& 50.2$^*$ & 49.4$^*$& 62.6$^*$ & 58.5 & 57.2$^*$ & 57.8$^*$ & 57.9$^*$ & 54.7 & 53.0 & 53.8& 62.5$^*$ & 62.6 & 63.9 & 62.8$^*$\\
        \text{SG-USM-L\&U} &\bf{52.3}$^*$&\bf{50.4}$^*$& \bf{51.4}$^*$ & \bf{50.9}$^*$ & \bf{64.7}$^*$ & \bf{61.6}$^*$ & \bf{58.8}$^*$ & \bf{60.2}$^*$ & \bf{58.4}$^*$ & \bf{55.8}$^*$ & \bf{53.2}$^*$ & \bf{54.5}$^*$ & \bf{63.3}$^*$ & \bf{63.1}$^*$ & \bf{64.1}$^*$ & \bf{63.5}$^*$\\
		\toprule
	\end{tabular}
}
\caption{Performance of SG-USM and baselines on various evaluation benchmarks. Numbers in \textbf{bold} denote the best model performance for a given metric. Numbers with $^*$ indicate that SG-USM model is better than the best-performing  baseline method (\underline{underlined} scores) with statistical significance (t-test, p < 0.05).}  
\label{tab:results}
\end{table*}

\subsection{Baselines and SG-USM Variants}
We compare our SG-USM approach with competitive baselines as well as state-of-the-art methods in user satisfaction modeling.

\noindent\textbf{HiGRU}~\citep{jiao2019higru}
 proposes a hierarchical structure to encode each turn in the dialogue using a word-level gated recurrent unit (GRU)~\citep{dey2017gate} and a sentence-level GRU. It uses the last hidden states of the sentence-level GRU as inputs of a multilayer perceptron (MLP)~\citep{hastie2009elements} to predict the user satisfaction level.
 
\noindent\textbf{HAN}~\citep{yang2016hierarchical}
applies a two-level attention mechanism in the hierarchical structure of HiGRU to represent dialogues. An MLP takes the dialogue representation as inputs to predict the user satisfaction level.

\noindent\textbf{Transformer}~\citep{vaswani2017attention}
is a simple baseline that takes the dialogue context as input and uses the standard Transformer encoder to obtain the dialogue representations. An MLP is used on the encoder to predict the user satisfaction level.

\noindent\textbf{BERT}~\citep{kenton2019bert}
concatenates the last 512 tokens of the dialogue context into a long sequence with a [SEP] token for separating dialogue turns. It uses the [CLS] token of a pre-trained BERT models to represent dialogues. An MLP is used on the BERT to predict the user satisfaction level.

\noindent\textbf{USDA}~\citep{deng2022user}
employs a hierarchical BERT encoder to encode the whole dialogue context at the turn-level and the dialogue-level. It also incorporates the sequential dynamics of dialogue acts with the dialogue context in a multi-task framework for user satisfaction modeling.

We also report the performance of two simpler SG-USM variants:

\noindent\textbf{SG-USM(L)}
only uses the dialogues with ground-truth user satisfaction labels to train the model.

\noindent\textbf{SG-USM(L\&U)}
uses both labeled and unlabeled dialogues in the training process. It takes the dialogues without user satisfaction annotation as the inputs of task attribute importance predictor module to obtain more general and accurate task attribute importance scores.

For a fair comparison with previous work and without loss of generality, we adopt BERT as the backbone encoder for all methods that use pre-trained language models.

\subsection{Evaluation Metrics}
Following previous work~\citep{deng2022user,cai2020predicting,choi2019offline,song2019using}, we consider a three-class classification task for user satisfaction modeling by treating the rating ``</=/> 3'' as ``dissatisfied/neutral/satisfied''. Accuracy (Acc), Precision (P), Recall (R), and F1 are used as the evaluation metrics.

\subsection{Training}
We use BERT-Base uncased, which has 12 hidden layers of 768 units and 12 self-attention heads to encode the utterances and schema descriptions. We apply a two-layer MLP with the hidden size as 768 on top of the text encoders. ReLU is used as the activation function. The dropout probability is 0.1. Adam~\citep{kingma2014adam} is used for optimization with an initial learning rate of 1e-4. We train up to 20 epochs with a batch size of 16, and select the best checkpoints based on the F1 score on the validation set. 

\section{Experimental Results}
\subsection{Overall Performance}
Table~\ref{tab:results} shows the results of SG-USM on MWOZ, SGD, ReDial, and JDDC datasets. Overall, we observe that SG-USM substantially and consistently outperforms all other methods across four datasets with a noticeable margin.  Specifically, SG-USM(L) improves the performance of user satisfaction modeling via explicitly modeling the degree to which the task attributes are fulfilled. SG-USM(L\&U) further aids the user satisfaction modeling via predicting task attribute importance based on both labeled dialogues and unlabeled dialogues. It appears that the success of SG-USM is due to its architecture design which consists of the task attribute fulfillment representation layer and the task attribute importance predictor. In addition, SG-USM can also effectively leverage unlabeled dialogues to alleviate the cost of user satisfaction score annotation.

\begin{figure*}[!h]
  \centering
  \subfloat[MWOZ]
    {
    \includegraphics[scale=0.22]{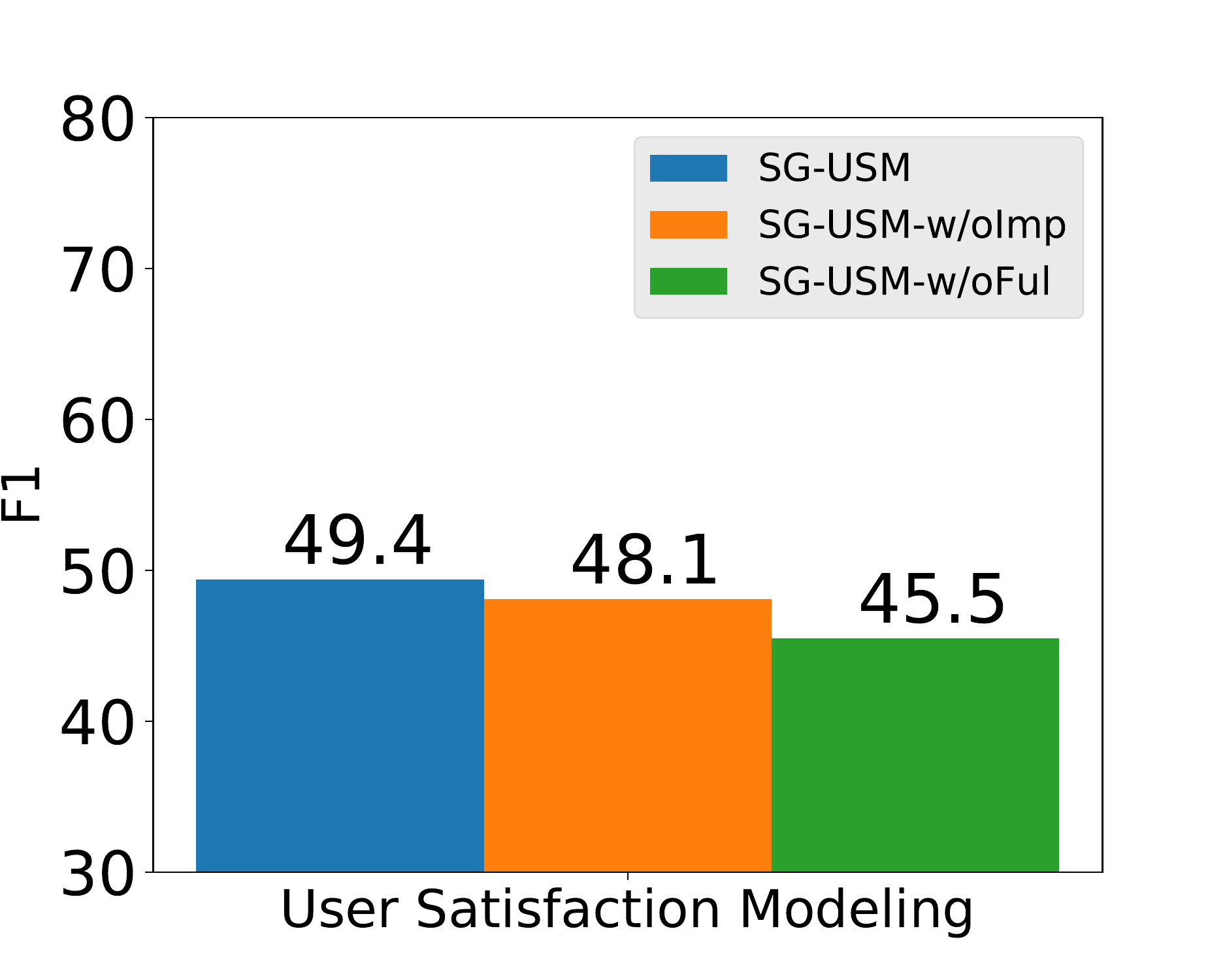}
    }
  \subfloat[SGD]
    {
    \includegraphics[scale=0.22]{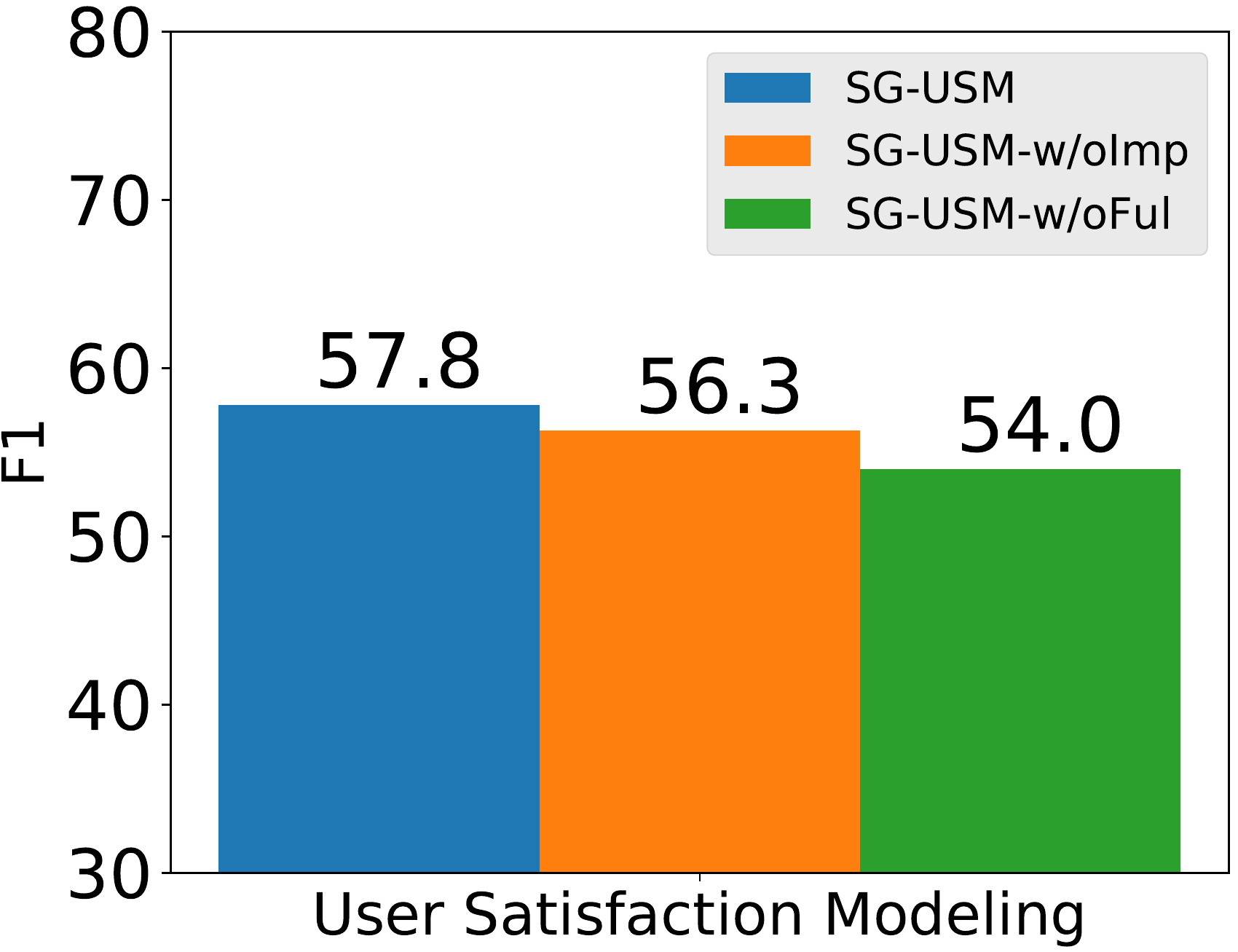}
    }
  \subfloat[ReDial]
    {
    \includegraphics[scale=0.22]{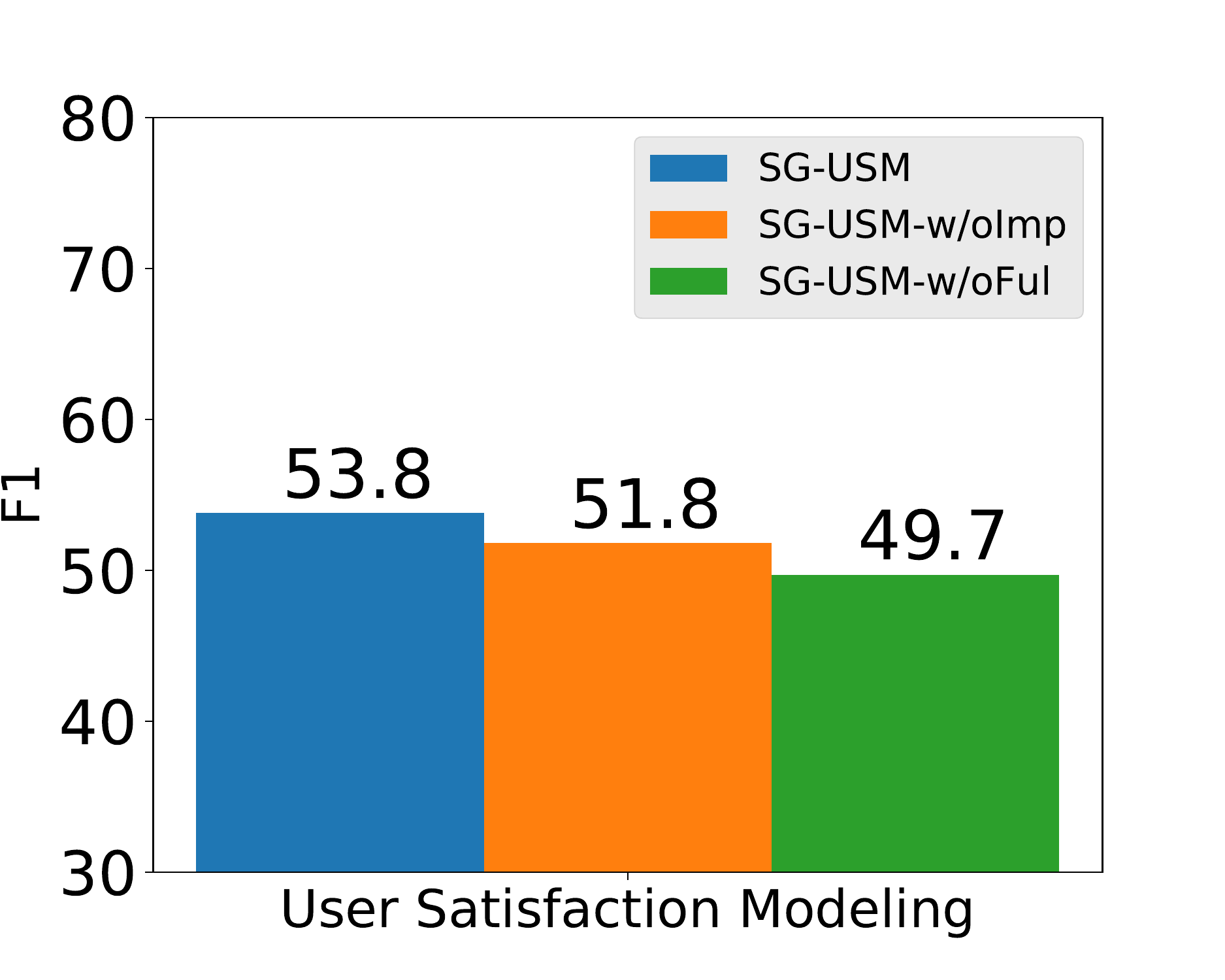}
    }
  \subfloat[JDDC]
    {
    \includegraphics[scale=0.22]{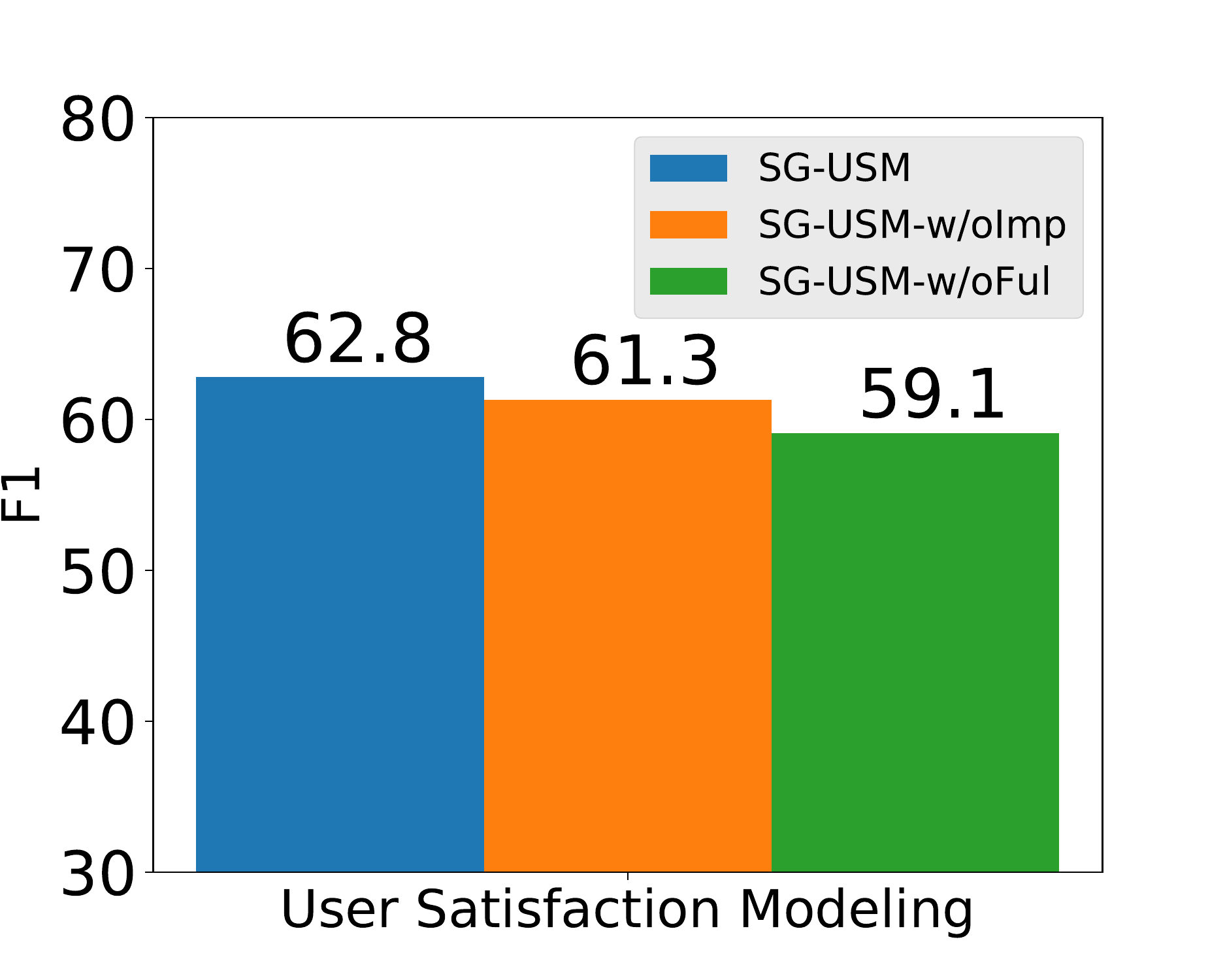}
    }
\caption{Performance of SG-USM by ablating the task attribute importance and task attribute fulfillment components across datasets.}
\label{fig:ablation}
\end{figure*}

\begin{figure*}[!h]
  \centering
  \subfloat[Example 1]
    {
    \includegraphics[scale=0.36]{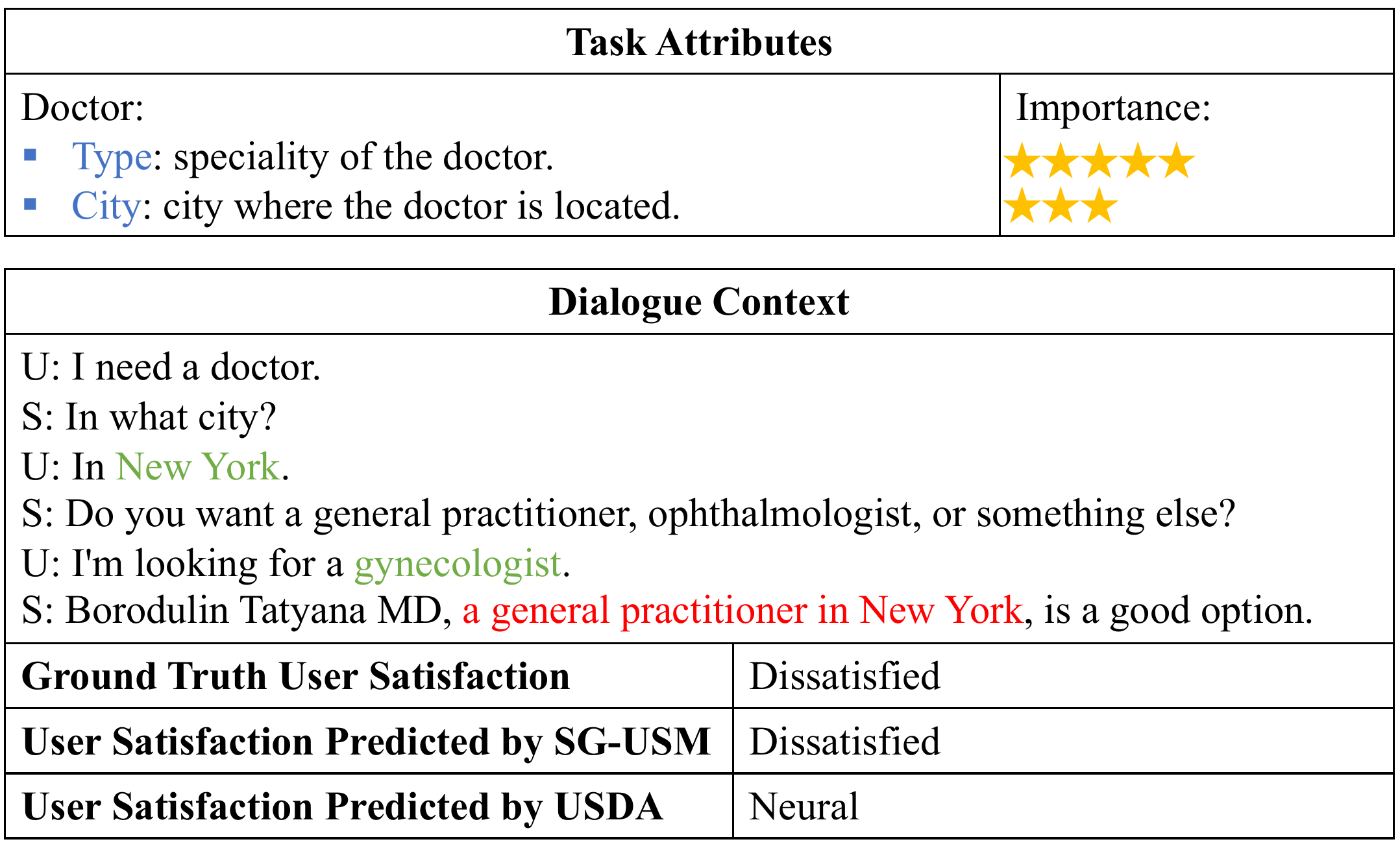}
    }
  \subfloat[Example 2]
    {
    \includegraphics[scale=0.36]{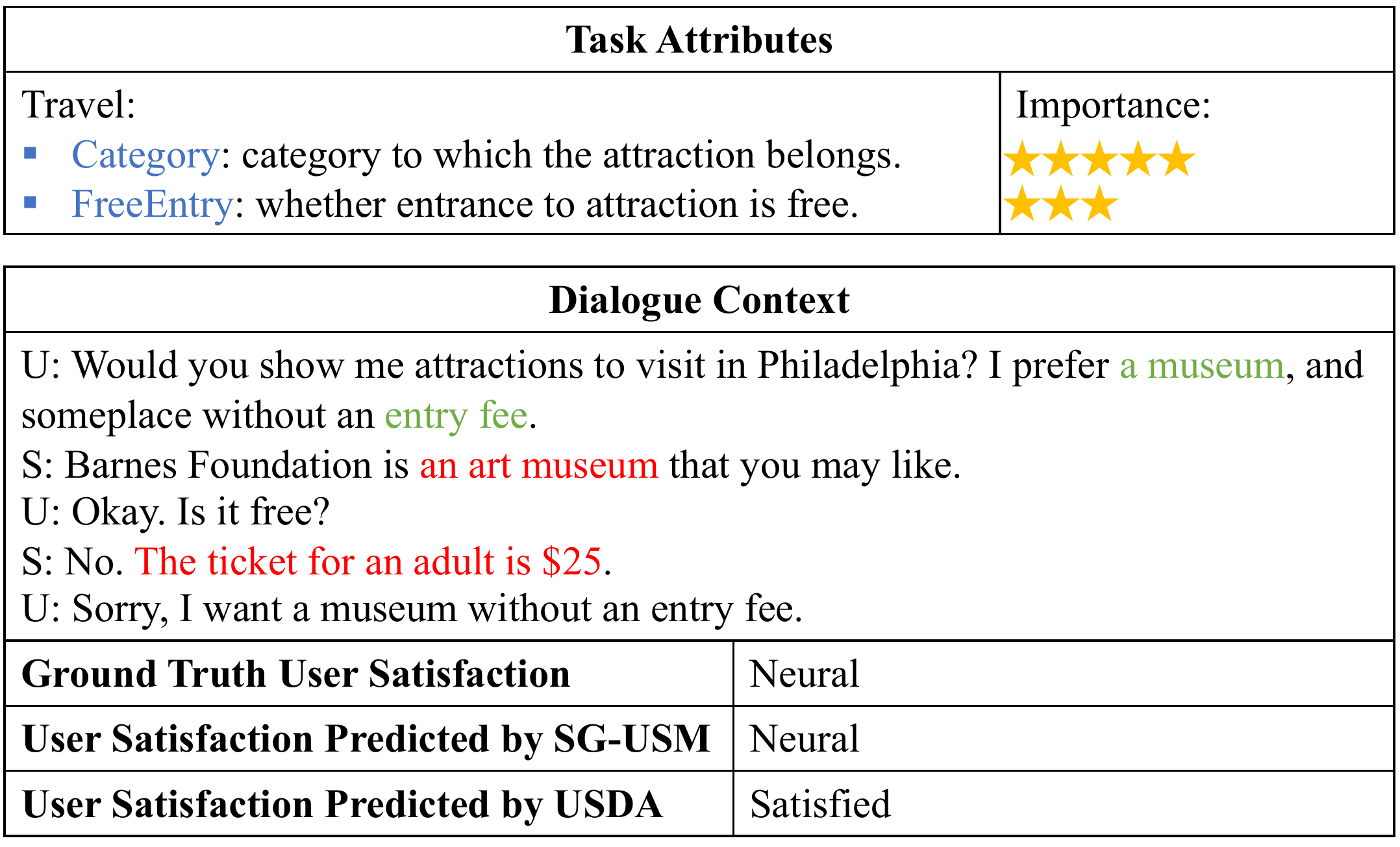}
    }
\caption{Case study on SG-USM and USDA on SGD dataset. The yellow $\filledstar$ represents the importance of task attributes. The texts in green are the users' preferences for the task attributes. The texts in red are the attributes of the provided solutions.}
\label{fig:case}
\end{figure*}

\subsection{Ablation Study}
We also conduct an ablation study on SG-USM to study the contribution of its two main  components: task attribute importance and task attribute fulfillment. 

\subsubsection*{Effect of Task Attribute Importance}
To investigate the effectiveness of task attribute importance in user satisfaction modeling, we eliminate the task attribute importance predictor and run the model on MWOZ, SGD, ReDial, and JDDC. As shown in Figure~\ref{fig:ablation}, the performance of SG-USM-w/oImp decreases substantially compared with SG-USM.
This indicates that the task attribute importance is essential for user satisfaction modeling. We conjecture that it is due to the user satisfaction relates to the importance of the fulfilled task attributes.

\subsubsection*{Effect of Task Attribute Fulfillment}
To investigate the effectiveness of task attribute fulfillment in user satisfaction modeling, we compare SG-USM with SG-USM-w/oFul which eliminates the task attribute fulfillment representation. Figure~\ref{fig:ablation} shows the results on MWOZ, SGD, ReDial, and JDDC in terms of F1. From the results, we can observe that without task attribute fulfillment representation the performances deteriorate considerably.
Thus, utilization of task attribute fulfillment representation is necessary for user satisfaction modeling.

\begin{table*}[!h]
\centering
\resizebox{0.75\textwidth}{!}{
\begin{tabular}{l|cccc|cccc}
        \toprule
        \multirow{2}{*}{{ \bf{Model}}}&\multicolumn{4}{c|}{{ \bf{MWOZ}}}&\multicolumn{4}{c}{{ \bf{ReDial}}}\\
        \cline{2-9}
        &{Acc}& {P} &{R}&{F1} &{Acc} & {P} &{R}  &{F1} \\
 		\hline
        \hline
        \text{USDA} &32.8 &34.5 & 32.2 & 33.1 & 25.4 & 29.5& 26.4& 27.3\\
        \text{SG-USM(L)} & 40.9$^*$ &38.9$^*$& 41.3$^*$&40.2$^*$ & 30.8$^*$ &34.6$^*$ & 30.7$^*$& 32.1$^*$\\
        \bf{\text{SG-USM(L\&U)}} &\bf{43.1}$^*$  &\bf{40.9}$^*$ & \bf{43.5}$^*$& \bf{42.8}$^*$ & \bf{32.3}$^*$ &\bf{36.4}$^*$ & \bf{32.8}$^*$& \bf{33.4}$^*$\\
		\toprule
	\end{tabular}
}
\caption{Performance of SG-USM and the best baseline USDA on zero-shot learning ability. All the models are trained on SGD and tested on MWOZ and ReDial. Numbers in \textbf{bold} denote best results in that metric. Numbers with $^*$ indicate that the model is better than the performance of baseline with statistical significance (t-test, p < 0.05).}  
\label{tab:unseen}
\end{table*}

\begin{figure}[!t]
\centering
\includegraphics[scale=0.72]{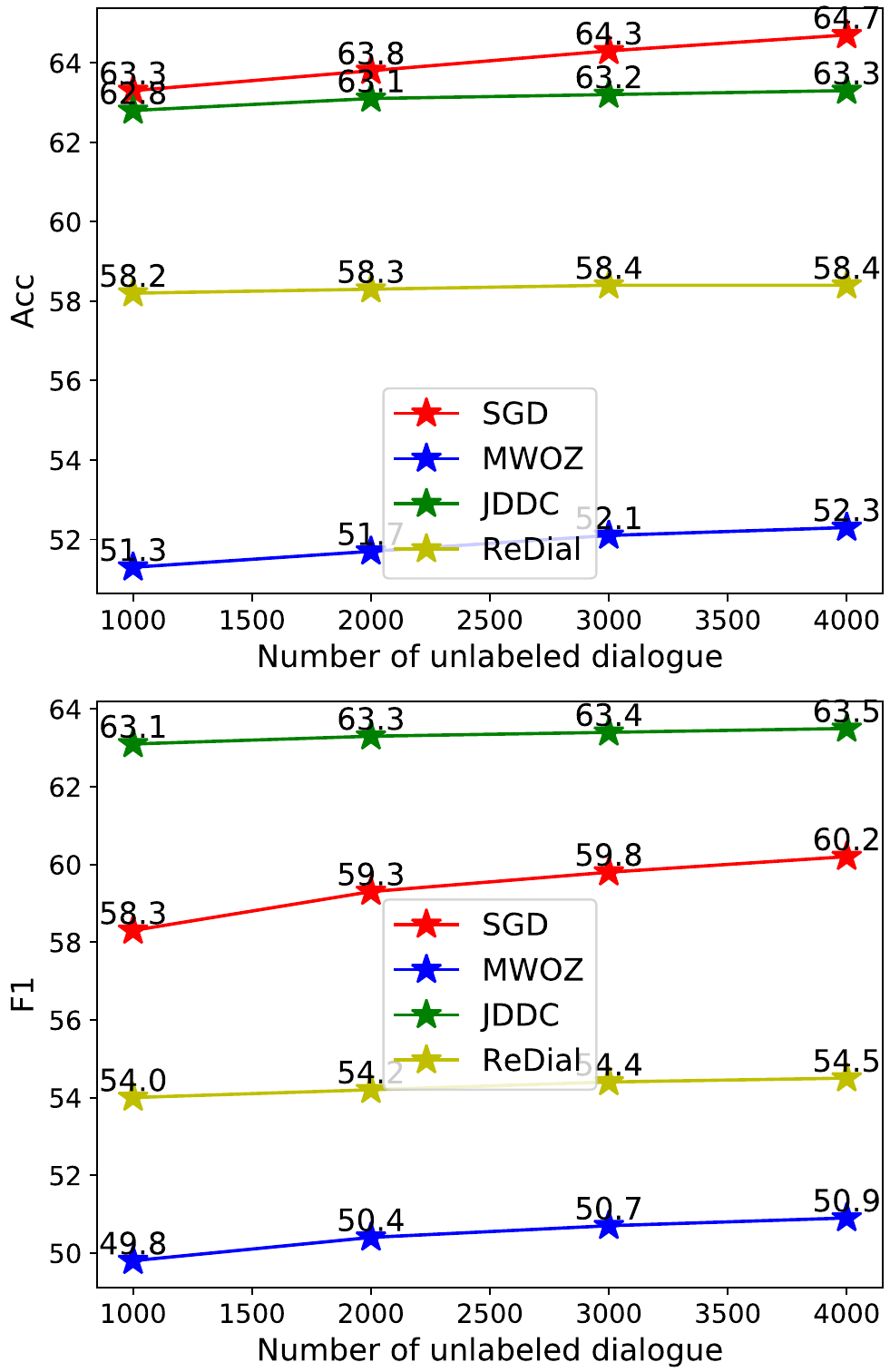}
\caption{Performance of SG-USM trained with different numbers of unlabeled dialogues on MWOZ, SGD, ReDial, and JDDC datasets.}
\label{fig:unlabel}
\end{figure}

\subsection{Discussion}

\subsection*{Case Study}
We also perform a qualitative analysis on the results of SG-USM and the best baseline USDA on the SGD dataset to delve deeper into the differences of the two models. 

We first find that SG-USM can make accurate inferences about user satisfaction by explicitly modeling the fulfillment degree of task attributes. For example, in the first case in Figure~\ref{fig:case}, the user wants to find a gynecologist in New York. SG-USM can correctly predict the dissatisfied label by inferring that the first important task attribute ``Type'' is not fulfilled. In the second case, the user wants to find a museum without an entry fee. SG-USM can yield the correct neural label by inferring that the second important task attribute ``FreeEntry'' is not fulfilled. From our analysis, we think that SG-USM achieves better accuracy due to its ability to explicitly model how many task attributes are fulfilled and how important the fulfilled task attributes are. In contrast, the USDA does not model the fulfillment degree of task attributes, thus it cannot properly infer the overall user satisfaction. 

\subsection*{Dealing with Unseen Task Attributes}
We furhter analyze the zero-shot capabilities of SG-USM and the best baseline of USDA. The SGD, MWOZ, and ReDial datasets are English dialogue datasets that contain different task attributes. Therefore, we train models on SGD, and test models on MWOZ and ReDial to evaluate the zero-shot learning ability. Table~\ref{tab:unseen} presents the Accuracy, Precision, Recall, and F1 of SG-USM and USDA on MWOZ and ReDial. From the results, we can observe that SG-USM performs significantly better than the baseline USDA on both datasets. This indicates that the agnostic task attribute encoder of SG-USM is effective. We argue that it can learn shared knowledge between task attributes and create more accurate semantic representations for unseen task attributes to improve performance in zero-shot learning settings.

\subsection*{Effect of the Unlabeled Dialogues}
To analyze the effect of the unlabeled dialogues for SG-USM, we test different numbers of unlabeled dialogues during the training process of SG-USM. Figure~\ref{fig:unlabel} shows the Accuracy and F1 of SG-USM when using 1 to 4 thousand unlabeled dialogues for training on MWOZ, SGD, ReDial, and JDDC. From the results, we can see that SG-USM can achieve higher performance with more unlabeled dialogues.
This indicates that SG-USM can effectively utilize unlabeled dialogues to improve the performance of user satisfaction modeling. We reason that with a larger corpus, the model can more accurately estimate the importance of task attributes. 

\section{Conclusion}
User satisfaction modeling is an important yet challenging problem for task-oriented dialogue systems evaluation. For this purpose, we proposed to explicitly model the degree to which the user's task goals are fulfilled. Our novel method, namely SG-USM, models user satisfaction as a function of the degree to which the attributes of the user's task goals are fulfilled, taking into account the importance of the attributes. Extensive experiments show that SG-USM significantly outperforms the state-of-the-art methods in user satisfaction modeling on various benchmark datasets, i.e. MWOZ, SGD, ReDial, and JDDC. Our extensive analysis also validates the benefit of explicitly modeling the fulfillment degree of a user's task goal based on the fulfillment of its constituent task attributes. 
In future work, it is worth exploring the reasons of user dissatisfaction to better evaluate and improve task-oriented dialogue systems.

\section*{Limitations}
Our approach builds on a task schema that characterizes a task-oriented dialogue system's domain. For example, the schema captures various attributes of the task. For some domains, when a schema is not pre-defined, it first needs to be extracted, e.g., from a corpus of dialogues. In this paper, we used BERT as our LM to be comparable with related work, but more advanced models could further improve the performance. A limitation of our task attribute importance scoring method is that it currently produces a static set of weights, reflecting the domain. In the future, the importance weights may be personalized to the current user's needs instead.


\bibliography{anthology}
\bibliographystyle{acl_natbib}




\end{document}